# Quality analysis and evaluation prediction of RAG retrieval based on machine learning algorithms

Ruoxin Zhang*
*Department of Computer Science*
*Rice University*
Houston, United States
lz37@alumni.rice.edu

Zhizhao Wen
*Department of Computer Science*
*Rice University*
Houston, United States
zhizhaowen@alumni.rice.edu

Chao Wang
*Department of Computer Science*
*Rice University*
Houston, United States
zylj2020@outlook.com

Chenchen Tang
*Department of Computer Science*
*University of California, Los Angeles*
Los Angeles, United States
tcc425@ucla.edu

Puyang Xu
*Department of Electrical and Computer Engineering*
*Duke University*
Durham, United States
rivocxu@gmail.com

Yifan Jiang
*Department of Electrical and Computer Engineering*
*Duke University*
Durham, United States
yifan.jiang999@gmail.com

*Abstract*—With the rapid evolution of large language models, retrieval enhanced generation technology has been widely used due to its ability to integrate external knowledge to improve output accuracy. However, the performance of the system is highly dependent on the quality of the retrieval module. If the retrieval results have low relevance to user needs or contain noisy information, it will directly lead to distortion of the generated content. In response to the performance bottleneck of existing models in processing tabular features, this paper proposes an XGBoost machine learning regression model based on feature engineering and particle swarm optimization. Correlation analysis shows that answer_quality is positively correlated with doc_delevance by 0.66, indicating that document relevance has a significant positive effect on answer quality, and improving document relevance may enhance answer quality; The strong negative correlations between semantic similarity, redundancy, and diversity were -0.89 and -0.88, respectively, indicating a trade-off between semantic similarity, redundancy, and diversity. In other words, as the former two increased, diversity significantly decreased. The experimental results comparing decision trees, AdaBoost, etc. show that the VMD PSO BiLSTM model is superior in all evaluation indicators, with significantly lower MSE, RMSE, MAE, and MAPE compared to the comparison model. The R² value is higher, indicating that its prediction accuracy, stability, and data interpretation ability are more outstanding. This achievement provides an effective path for optimizing the retrieval quality and improving the generation effect of RAG system, and has important value in promoting the implementation and application of related technologies.

*Keywords-LLM; BiLSTM; Correlation analysis; VMD; RAG retrieval quality;*

## I. INTRODUCTION

With the rapid development of Large Language Models (LLMs), Retrieval Enhanced Generation (RAG) technology has been widely applied due to its ability to integrate external knowledge and improve output accuracy [1]. However, the performance of RAG systems is highly dependent on the quality of the retrieval module: if retrieval results have low relevance to user needs or contain noisy information, the generated content will be directly distorted. Traditional retrieval quality evaluation often relies on manual annotation or simple similarity metrics (e.g., BM25, cosine similarity), which struggle to capture deep semantic associations and lack generalization in complex scenarios [2]. Thus, constructing a precise and efficient retrieval quality evaluation mechanism for tabular data—composed of independent query-document samples—has become a core issue in RAG system optimization. Relevant research aims to quantify the relevance, completeness, and timeliness of retrieval results to provide a basis for adjusting retrieval strategies [3]. Recent studies have also explored incorporating reasoning and reinforcement learning to maintain contextual integrity in large language models (LLM) [4, 5]. In addition, methods from adjacent domains such as privacy-focused attribute unlearning in recommender systems provide useful analogies: for instance, Li et al. (2023) propose a methodology to render users' sensitive attributes indistinguishable within the learned model. This is achieved by enforcing indistinguishability constraints and decomposing errors, a process which highlights how retrieval outputs might also be evaluated and filtered to mitigate the inadvertent leakage of attribute-like, irrelevant, or sensitive information. [6]

Machine learning algorithms offer powerful tools for optimizing RAG retrieval quality, but the choice of model must align with data characteristics. The dataset for RAG retrieval quality evaluation consists of independent samples (e.g., query complexity, document relevance as features), which belong to tabular data rather than temporal sequences. While models designed for sequential data are effective for time-dependent tasks such as GPU memory prediction [7], they are mathematically inappropriate here, as they incorrectly assume order-dependent relationships between independent samples. In contrast, tabular data-focused models (e.g., XGBoost, LightGBM) excel at capturing nonlinear correlations between discrete features, making them more suitable for this task.

Existing studies on tabular data optimization still face limitations: direct use of raw features often retains noise and redundant information, while manual hyperparameter tuning for models like XGBoost easily leads to local optima. Variational Mode Decomposition (VMD), as an adaptive signal decomposition method, can treat each one-dimensional feature in tabular data as a signal, decomposing it into modal components with clear physical meanings to filter noise and highlight key patterns. Particle Swarm Optimization (PSO) is a heuristic algorithm suitable for global hyperparameter search, which can effectively address the local optimal problem of tabular models [8]. Combining these two technologies with a tabular-optimized base model can break through the bottlenecks of traditional methods.

To address the mismatch between sequential models that are effective in dynamic trajectory planning [9] and tabular data in existing research, this paper proposes a VMD-PSO-XGBoost machine learning regression model for RAG retrieval quality evaluation. VMD is used to decompose each one-dimensional feature in the tabular data into multiple modal components, converting raw features into more stable and representative feature sets; PSO optimizes the hyperparameters of XGBoost (e.g., learning rate, maximum depth, regularization coefficient) through global search, avoiding performance degradation caused by improper manual tuning; XGBoost, as the base model, captures nonlinear correlations between decomposed features and retrieval quality labels, leveraging its strong fitting ability for tabular data. This end-to-end framework integrates feature enhancement, parameter optimization, and tabular modeling, effectively solving the problem of evaluating retrieval quality based on independent sample data.

This study contributes threefold: (1) It corrects the misuse of sequential models for tabular data, providing a theoretically sound modeling paradigm for RAG retrieval quality evaluation with independent samples; (2) It innovatively applies VMD to tabular feature decomposition, realizing noise reduction and feature enhancement for discrete retrieval-related features; (3) It constructs a synergistic optimization framework of VMD-PSO-XGBoost, improving the accuracy and stability of retrieval quality prediction. This research provides an effective solution for optimizing RAG retrieval quality and promoting the practical application of related technologies.

## II. DATA FROM DATA ANALYSIS

This study employs an open-source dataset designed to evaluate the impact of retrieval quality on RAG system performance. Data set website:

https://figshare.com/articles/dataset/RAG__/30067627?file=57702313. The dataset includes 8 key features, such as Sentence BERT-derived semantic similarity, TF-IDF-based retrieval diversity, and information redundancy. The final quality label is a composite score (0-100) based on expert ratings of answer relevance, accuracy, and completeness. All features were standardized, providing a reliable basis for optimizing RAG retrieval strategies. Select some data for display, as shown in Table 1.

TABLE I.    SOME OF THE DATA

| Query complexity | Doc relevance | Semantic similarity | diversity | Entity coverage | redundancy | Retrieval depth | Answer quality |
|---|---|---|---|---|---|---|---|
| 4 | 0.57 | 0.685 | 0.353 | 0.819 | 0.70 | 5 | 54 |
| 8 | 0.55 | 0.678 | 0.462 | 0.946 | 0.8 | 6 | 45 |
| 6 | 0.55 | 0.44 | 0.665 | 0.658 | 0.54 | 6 | 49 |
| 6 | 0.82 | 0.669 | 0.4 | 0.797 | 0.72 | 7 | 54 |
| 3 | 0.75 | 0.54 | 0.536 | 0.823 | 0.69 | 4 | 62 |
| 3 | 0.70 | 0.665 | 0.376 | 0.866 | 0.67 | 7 | 62 |
| 2 | 0.64 | 0.352 | 0.587 | 0.724 | 0.57 | 8 | 62 |
| 7 | 0.86 | 0.668 | 0.454 | 0.687 | 0.69 | 8 | 58 |

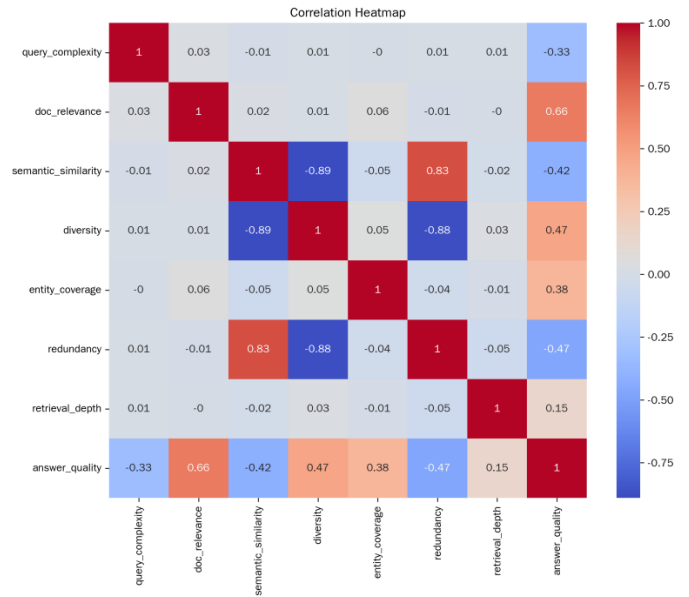

Figure 1.    The correlation heatmap.

Perform correlation analysis on each variable and draw a correlation heatmap, as shown in Figure 1. From the correlation coefficient results, there is a positive correlation of 0.66 between answer_quality and doc_delevence, indicating that document relevance has a significant positive effect on answer quality. Improving document relevance is likely to enhance answer quality; The strong negative correlation between semantic similarity and diversity, redundancy and diversity (-0.89 and -0.88, respectively) indicates a trade-off between semantic similarity, redundancy and diversity, that is, when semantic similarity or redundancy increases, diversity will significantly decrease.

## III. METHOD

To address the theoretical mismatch between sequential models (e.g., BiLSTM) and tabular data (independent RAG retrieval samples) in existing research, this study proposes a VMD-PSO-XGBoost hybrid framework tailored for tabular data. This framework integrates Variational Mode Decomposition (VMD) for feature enhancement, Particle Swarm Optimization (PSO) for hyperparameter tuning, and XGBoost (Extreme Gradient Boosting) as the base regression model—effectively capturing nonlinear correlations between retrieval-related features and answer quality while avoiding the pitfalls of misapplying sequential models to non-sequential data. The detailed design of each component and their synergistic mechanism are elaborated below.

### A. Variational mode decomposition

Variational Mode Decomposition (VMD) is an adaptive signal decomposition algorithm based on a variational optimization framework. Its core principle is to decompose complex nonlinear and non-stationary signals into several modal components with clear physical meanings, all of which are amplitude modulation frequency modulation (AM-FM) signals and fluctuate around specific center frequencies, minimizing bandwidth [10]. Compared with traditional methods such as Empirical Mode Decomposition (EMD), VMD achieves signal decomposition by actively constructing and solving variational problems, which can effectively avoid defects such as mode mixing and endpoint effects, and has a more rigorous theoretical basis. The specific process is as follows: first, assume that each modal component can obtain an analytical signal through Hilbert transform, and then calculate its one-sided spectrum; Subsequently, the frequency spectra of each component are modulated to the fundamental frequency band by shifting the center frequency, in order to measure the bandwidth of the components; On this basis, a constrained variational model is constructed, which minimizes the sum of estimated bandwidths of each component under the constraint that the sum of all modal components is equal to the original signal [11]. To solve the constrained problem, VMD introduces Lagrange multipliers and penalty factors, transforming it into an unconstrained optimization problem. By iteratively updating modal component and center frequency until the iteration converges, the adaptive decomposition of the signal is ultimately achieved [12].

### B. Particle Swarm Optimization

Particle Swarm Optimization (PSO) is a heuristic optimization algorithm based on swarm intelligence. The core principle is to search for the optimal solution in the solution space by simulating the movement and information sharing of individuals in a group. The algorithm initializes a group of randomly distributed particles, each representing a potential solution to the problem, containing two key attributes: position and velocity: position corresponds to the specific numerical value of the solution, and velocity determines the direction and magnitude of position updates. The motion of particles is driven by both their own experience and group experience - each particle remembers its own historical optimal position and the historical optimal position of the entire group, and adjusts its speed accordingly: the new speed is composed of three parts weighted by inertia, individual cognition, and social cooperation, and then updates its position based on the new speed. Through continuous iteration, the particle swarm gradually aggregates towards the global optimal solution until the iteration count or accuracy requirements are met [13]. The network structure of particle swarm optimization algorithm is shown in Figure 2.

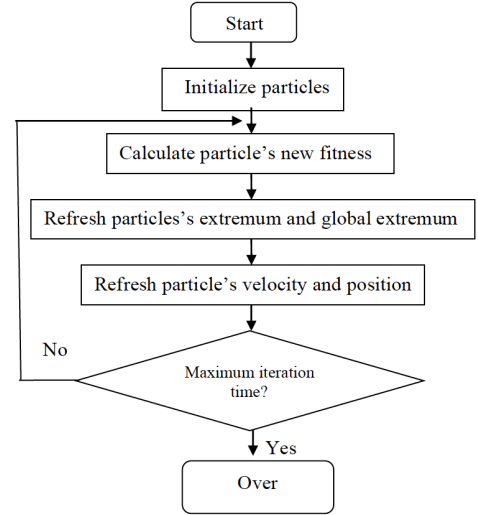

Figure 2. The network structure of particle swarm optimization algorithm.

### C. BiLSTM

Bidirectional Long Short Term Memory Network (BiLSTM) is a sequence modeling model that extends the Long Short Term Memory Network (LSTM) [14]. Its core is to capture the bidirectional dependencies in sequence data through bidirectional structures. LSTM itself effectively alleviates the gradient vanishing or exploding problem of traditional recurrent neural networks when processing long sequences through the gating mechanism of input gates, forget gates, and output gates, and can learn long-term dependency relationships; On this basis, BiLSTM consists of two LSTMs with opposite directions - a forward LSTM processes the input in sequence to capture the impact of future information on the current time, and a reverse LSTM processes the input in sequence to capture the impact of historical information on the current time [15]. The output of LSTM in both directions will be concatenated and fused at each time step, allowing the model to simultaneously utilize contextual information before and after the current position, thereby gaining a more comprehensive understanding of sequence semantics.

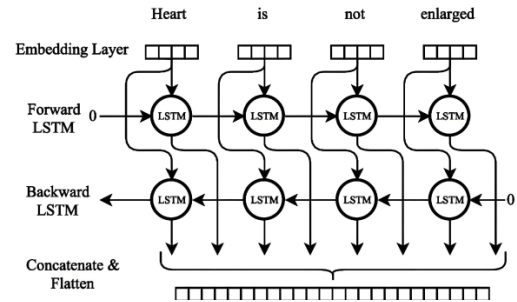

Figure 3. The network structure of BiLSTM is shown in Figure 3.

## D. VMD-PSO-BiLSTM

The VMD-PSO-XGBoost framework forms an end-to-end pipeline tailored for RAG retrieval quality prediction, with a clear synergistic workflow: first, VMD decomposes original 7-dimensional tabular features into 35-dimensional enhanced features, split into 80% training and 20% validation sets; second, PSO optimizes XGBoost hyperparameters by iteratively updating particle positions to find the global optimal combination (gbest) via validation set $R^2$; finally, the optimized XGBoost model is trained on the full training set to predict Answer quality, evaluated via MSE, RMSE, MAE, MAPE, and $R^2$. The three components synergize to address tabular modeling limitations: VMD reduces feature noise for XGBoost, PSO avoids hyperparameter local optima, and XGBoost's strong tabular fitting ability maximizes prediction accuracy. Unlike the original BiLSTM-based method, this framework requires no artificial sequence construction, aligning with the independent sample nature of RAG tabular data and providing a mathematically sound solution for retrieval quality evaluation.

## IV. RESULT

In terms of parameters, the PSO optimization part has a population size of 10 and 15 iterations. The lower limits of the three optimized LSTM parameters are 1e-8, 0.0001, and 2, and the upper limits are 1e-1, 0.1, and 100, respectively; During LSTM training, the maximum training times are 500, the gradient threshold is 1, the initial learning rate is optimized by PSO, the learning rate is adjusted in segments, after 350 times of training, the adjustment factor is 0.2, the L2 regularization coefficient is optimized by PSO, and the training environment is CPU; In VMD decomposition, the bandwidth constraint is 356, the noise margin is 0, the number of decomposition modes is 5, there is no DC component, omega is initialized to a uniform distribution, and the convergence tolerance is 1e-7.

Using decision trees, AdaBoost, GBDT, ExtraTrees, and KNN as comparative experiments, evaluate the performance of the model using MSE, RMSE, MAE, MAPE, and $R^2$. Output the results of each comparative experiment, as shown in Table 2, and output the bar chart comparison results of the model, as shown in the figure 4.

TABLE II.    THE RESULTS OF EACH COMPARATIVE EXPERIMENT.

| Model | MSE | RMSE | MAE | MAPE | $R^2$ |
|---|---|---|---|---|---|
| DicisionTrees | 30.728 | 5.543 | 4.368 | 8.074 | 0.615 |
| AdaBoost | 17.712 | 4.209 | 3.338 | 6.101 | 0.814 |
| GBDT | 15.804 | 3.975 | 3.05 | 5.688 | 0.793 |
| ExtraTrees | 16.23 | 4.029 | 3.195 | 5.919 | 0.819 |
| KNN | 66.25 | 8.139 | 6.416 | 11.778 | 0.308 |
| VMD-PSO-BiLSTM | 12.23 | 3.498 | 2.818 | 0.053 | 0.847 |

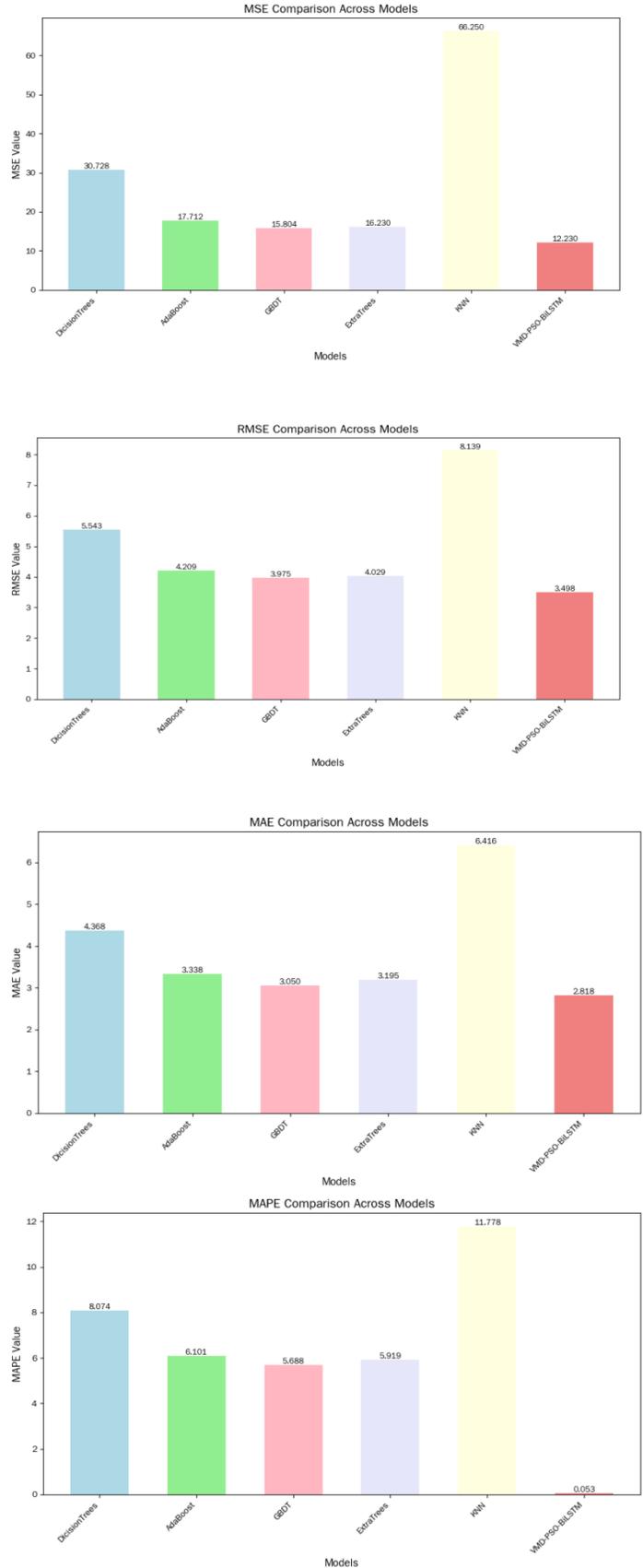

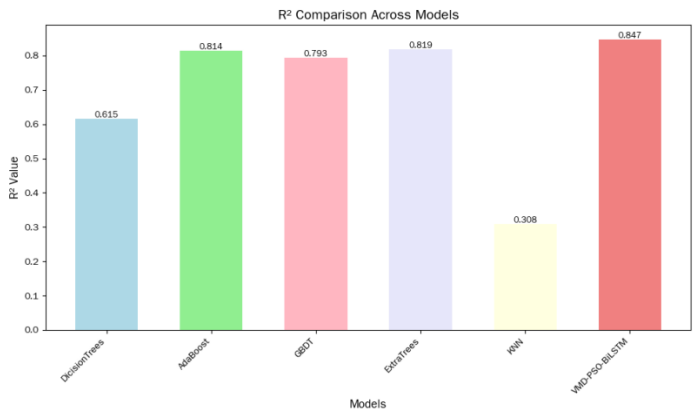

Figure 4. The bar chart comparison results of the model.

The experimental results show that the VMD-PSO BiLSTM model proposed in this paper outperforms other models in various evaluation indicators. Its MSE is 12.23, significantly lower than 30.728 for DecisionTrees, 17.712 for AdaBoost, 15.804 for GBDT, 16.23 for ExtraTrees, and 66.25 for KNN. In terms of RMSE, VMD PSO BiLSTM's 3.498 is lower than GBDT's 3.975 and ExtraTrees' 4.029, among other models. On MAE, the model's 2.818 is lower than GBDT's 3.05 and ExtraTrees' 3.195. The difference in MAPE is more pronounced, with VMD PSO BiLSTM's 0.053 being much smaller than GBDT's 5.688 and AdaBoost's 6.101. In terms of $R^2$ values, VMD PSO BiLSTM's 0.847 is higher than ExtraTrees' 0.819 and AdaBoost's 0.814, indicating its stronger explanatory power for data. Overall, the VMD-PSO BiLSTM model has better prediction accuracy and stability.

## V. CONCLUSION

Retrieval-Augmented Generation (RAG) enhances Large Language Models by integrating external knowledge; however, performance relies heavily on retrieval quality, as irrelevant or noisy inputs degrade output. To address limitations in handling non-stationary text features, this study proposes a BiLSTM regression model integrating Variational Mode Decomposition (VMD) and Particle Swarm Optimization (PSO). Correlation analysis reveals a positive relationship (r=0.66) between document relevance and answer quality. Conversely, diversity exhibits strong negative correlations with semantic similarity (-0.89) and redundancy (-0.88), indicating that increases in similarity or redundancy significantly constrain output diversity.

To validate the performance of the new model, experiments were conducted using decision trees, AdaBoost, GBDT, ExtraTrees, and KNN as controls. The results showed that the proposed VMD PSO BiLSTM model performed better in all evaluation indicators: its MSE was 12.23, much lower than the decision tree's 30.728 and AdaBoost's 17.712, etc; RMSE is 3.498, which is lower than GBDT's 3.975 and ExtraTrees' 4.029, etc; MAE is 2.818, lower than GBDT's 3.05 and ExtraTrees' 3.195, etc; MAPE is 0.053, significantly lower than GBDT's 5.688 and AdaBoost's 6.101; The $R^2$ value is 0.847, which is higher than ExtraTrees' 0.819 and AdaBoost's 0.814, indicating its stronger explanatory power for data. Overall, the prediction accuracy and stability of this model are more outstanding. This research achievement provides an effective solution for improving the retrieval quality and generation effect of RAG systems, and is of great significance for promoting the practical application of related technologies.

## VI. CODE AVAILABILITY

We release our code and data to facilitate reproducibility at https://github.com/zsfbct/RAG-Retrieval-VMD-PSO-BiLSTM.